\documentclass[10pt,twocolumn,letterpaper]{article}

\usepackage{iccv}
\usepackage{times}
\usepackage{epsfig}
\usepackage{graphicx}
\usepackage{amsmath}
\usepackage{amssymb}
\usepackage{verbatim}

\newcommand\norm[1]{\left\lVert#1\right\rVert}

\DeclareMathOperator*{\argmax}{arg\,max}
\DeclareMathOperator*{\argmin}{arg\,min}

\newcommand{\rulesep}{\unskip\ \hrule\ }

\usepackage{lipsum}
\usepackage{booktabs}
\usepackage{amsfonts}
\usepackage{standalone}
\usepackage{bm}
\usepackage{algorithm}
\usepackage{algpseudocode}
\usepackage{cite}
\usepackage{makecell}
\usepackage{courier}
\usepackage{subfig}
\usepackage{multirow}
\usepackage[table]{xcolor}
\usepackage{amsmath,amssymb}
\usepackage{amsthm}
\usepackage{gensymb}
\usepackage{xcolor}

\usepackage[pagebackref=true,breaklinks=true,letterpaper=true,colorlinks,bookmarks=false]{hyperref}

\iccvfinalcopy 

\def\iccvPaperID{3734} 

\begin{document}

\title{EMPNet: Neural Localisation and Mapping Using Embedded Memory Points}
\author{Gil Avraham
\and
Yan Zuo
\and
Thanuja Dharmasiri
\and
Tom Drummond \\
ARC Centre of Excellence for Robotic Vision, Monash University, Australia \\
{\tt\small \{gil.avraham, yan.zuo, thanuja.dharmasiri, tom.drummond\}@monash.edu}
}

\twocolumn[{
\renewcommand\twocolumn[1][]{#1}%
\maketitle
\begin{center}
\includegraphics[width=0.85\textwidth]{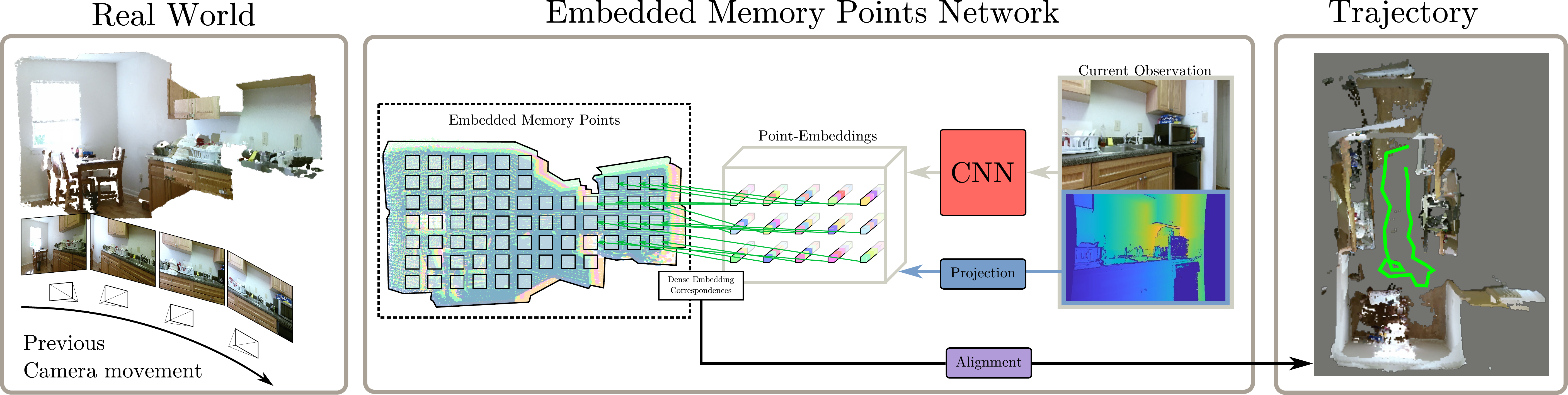}
\captionof{figure}{EMP-Net maintains an internal representation which corresponds to a real world environment. This internal spatial memory is continuously updated through a dense matching algorithm, allowing an autonomous agent to localise and model the world through sequences of observations.}
\label{fig:frontpage_ssmm}
\end{center}
}]

\begin{abstract}
Continuously estimating an agent's state space and a representation of its surroundings has proven vital towards full autonomy. A shared common ground among systems which successfully achieve this feat is the integration of previously encountered observations into the current state being estimated. This necessitates the use of a memory module for incorporating previously visited states whilst simultaneously offering an internal representation of the observed environment. In this work we develop a memory module which contains rigidly aligned point-embeddings that represent a coherent scene structure acquired from an RGB-D sequence of observations. The  point-embeddings are extracted using modern convolutional neural network architectures, and alignment is performed by computing a dense correspondence matrix between a new observation and the current embeddings residing in the memory module. The whole framework is end-to-end trainable, resulting in a recurrent joint optimisation of the point-embeddings contained in the memory. This process amplifies the shared information across states, providing increased robustness and accuracy. We show significant improvement of our method across a set of experiments performed on the synthetic VIZDoom environment and a real world Active Vision Dataset.
\end{abstract}
\footnotetext{This work was supported by the Australian Research Council Centre of Excellence for Robotic Vision (project number CE1401000016)}
\section{Introduction}

In recent times, there has been a large surge in interest towards developing agents which are fully autonomous. A core aspect of full autonomy lies in the spatial awareness of an agent about its surrounding environment~\cite{davison2003real}; this understanding would enable the extension towards other useful applications including navigation~\cite{davison2007monoslam} as well as human-robot interaction~\cite{corke1993visual}. Although performance in image understanding challenges such as segmentation~\cite{long2015fully,he2017mask,chen2018deeplab}, depth estimation~\cite{eigen2014depth,weerasekera2018just,spek2018cream}, video prediction~\cite{lotter2016deep,zuo2018traversing}, object classification~\cite{krizhevsky2012imagenet,simonyan2014very,he2016deep} and detection~\cite{girshick2015fast,ren2015faster} has seen vast improvement with the aid of deep learning, this level of success has yet to translate towards the intersection between spatial awareness and scene understanding. Currently, this is an active area of research~\cite{henriques2018mapnet, gupta2017cognitive, parisotto2017neural}, with the vision community realising its potential towards merging intelligent agents seamlessly and safely into real world environments.




Fundamentally, an autonomous agent is required to maintain an internal representation of the observed scene structure that may be accessed for performing tasks such as navigation, planning, object interaction and manipulation~\cite{gupta2017cognitive}. Traditional SLAM methods~\cite{davison2003real,mur2015orb} maintain an internal representation via keyframes, stored in a graph-like structure which provides an efficient approach for large-scale navigation tasks. Although, in order to distil the local structural information of a scene, dense representations~\cite{newcombe2011dtam} are often better suited for the given task. Incidentally, this dense representation is also more applicable for modern deep learning approaches.

Given this, we identify an important relationship between the representation of a scene structure and the geometric formulation of the problem. Nowadays, the increased popularity of cameras with depth sensors mounted on robotic platforms means that RGB-D information of the scene is readily available. For an agent navigating an environment whilst simultaneously collecting colour and depth information, a natural representation is a 3D point entity which can capture the spatial neighbourhood information of its surroundings. The alignment of such representations has been heavily explored in the literature~\cite{tam2013registration}.


In this work, we reformulate the task of finding 3D point correspondences as a cross-entropy optimisation problem. By having access to depth sensory and the agent's pose information in the data collection stage, a ground-truth correspondence matrix can be constructed between two consecutive frames such that 3D points which match are assigned a probability of `1', and non-matches assigned a `0'. Using a Convolutional Neural Network (CNN), we extract feature embeddings from an acquired observation, which are then assigned to projected depth points. Collectively, we refer to these embedding-coordinate pairs as point-embeddings. This allows for end-to-end optimisation on the correspondences between closest point-embeddings (Fig.~\ref{fig:frontpage_ssmm}). 




By iteratively repeating this process, extracted point embeddings stored from previously seen observations are jointly optimised within a processed sequence of frames, forming a recurrent memory mechanism. The point-embeddings along with their 3D location are stored within a memory component which we refer to as the Short-term Spatial Memory Module (SSMM). Through continuously inferring a correspondence matrix between point-embeddings in the SSMM and newly extracted point-embeddings, we obtain the relative pose between the incoming frame and a local coordinate frame of the SSMM. The result is a SSMM which contains point-embeddings which are structurally aligned to their original structure in the real world. 

We evaluate our method on two datasets: a synthetic environment from the Doom video-game, and a real environment captured from a robotic platform from the Active Vision Dataset. In both datasets, we show that our method significantly outperforms baselines on localisation tasks.

The rest of this paper is organised as follows: in Section~\ref{sec:related_work}, we give a brief review of related work. In Section~\ref{sec:empnet}, we provide details of our proposed method. In Section~\ref{sec:experiments}, we show experimental results and discuss possible extensions to our method in Section~\ref{sec:future_work}.

\section{Related Work}
\label{sec:related_work}
The related literature to our work can be organised into three categories.
\paragraph{Frame-based}
Prior to the introduction of memory based models for localisation and mapping tasks, frame-by-frame methods~\cite{kendall2015posenet,costante2016exploring} and more recently~\cite{dharmasiri2018eng, mahjourian2018unsupervised}, explored the exploitation of geometric constraints for reducing the search space when optimising Convolutional Neural Networks (CNN). The pioneering work of~\cite{kendall2015posenet} applied direct pose regression for inferring the relative pose between two views. The work by~\cite{costante2016exploring} enhanced the information provided to the regression network by including the optical flow between two consecutive frames. A natural extension was explored by~\cite{dharmasiri2018eng} which simultaneously estimated a depth map along with a latent optical flow constraint for regressing the pose between consecutive frames. CodeSLAM~\cite{bloesch2018codeslam} optimises an encoder-decoder setup to efficiently represent depth frames as latent codes. These latent codes are optimised so that pose information can be used to transform one latent code to another. Most recently,~\cite{mahjourian2018unsupervised} combined a photometric loss with depth estimation and additionally used the inferred depth for minimising the residual error of 3D iterative closest point~\cite{hartley2003multiple} loss. In our work, we similarly minimise a closest point loss, though we minimise the direct closest point errors between an internally modelled environment and an incoming observation.

\paragraph{Sequence-based}
The importance of maintaining an internal representation of previously seen observations was initially explored in DeepVO~\cite{wang2017deepvo} and VINet~\cite{clark2017vinet}. Both works process sequential information by extracting features using a CNN, which are inputted to an LSTM~\cite{hochreiter1997long} for fusing past observations while regressing the relative pose between two consecutive frames. DeepTAM~\cite{zhou2018deeptam} reformulated the Dense Tracking and Mapping (DTAM) approach of~\cite{newcombe2011dtam} as a learning problem. Similar to DeepVO, DeepTAM regresses the pose directly and in addition, estimates an expensive cost volume for mapping the environment. An elegant extension to the above approaches by~\cite{clark2017vidloc} exploits bidirectional LSTMs for obtaining temporally smooth pose estimates (however this bidirectional property introduces an inference lag). Similarly, we maintain a consistent spatial temporal representation of the environment, although our short-term memory recall is more verbose and engineered to have more explicit meaning of localising against previously seen observations.

\begin{figure*}[t]
\centering
\includegraphics[width=0.85\textwidth]{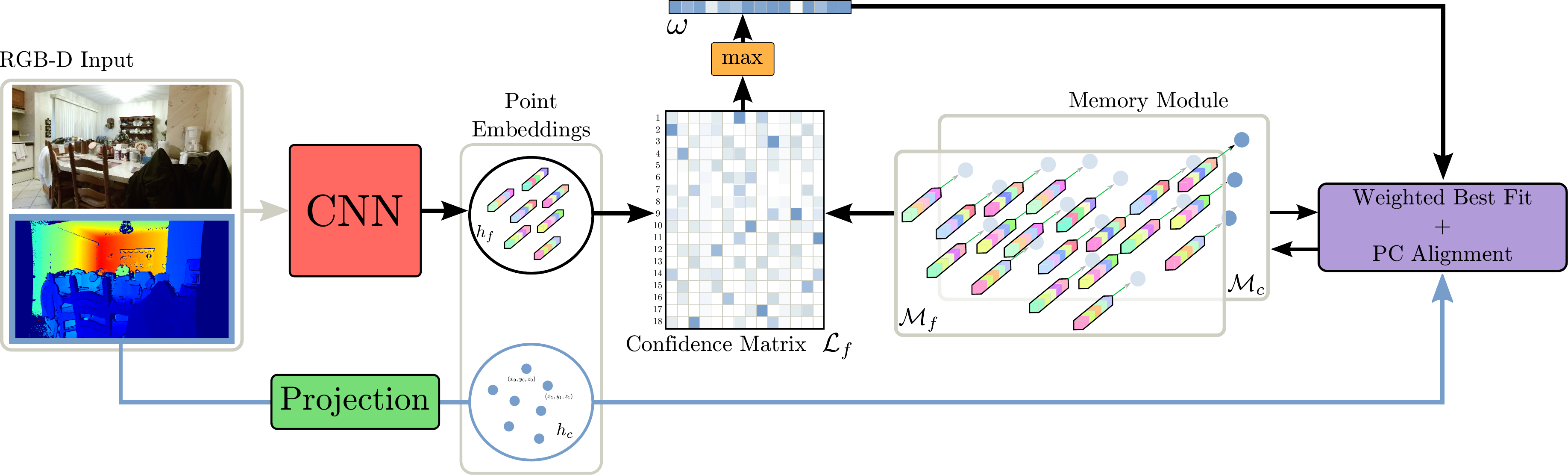}
\caption{The proposed architecture of EMP-Net. Incoming observations are processed to extract point-embeddings and localise against a short-term memory module. The memory module is updated with the new set of point-embeddings after aligning them to the memory's coordinate frame using the inferred pose.}
\label{fig:system_overview}
\end{figure*}

\paragraph{Map-based}
Incorporating a more explicit map representation of an agents previously visited states was explored by Reinforcement Learning based approaches~\cite{zhang2017neural,parisotto2017neural,gupta2017cognitive} where optimising towards a goal which forces an agent to model the environment was found beneficial. Both Neural SLAM~\cite{zhang2017neural} and Neural Map~\cite{parisotto2017neural} have a fixed latent map size, with a 2D top-down map representation. However, both of these works only assess their models on synthetic mazes and toy tasks.~\cite{gupta2017cognitive} extended upon this with the introduction the Cognitive Mapper and Planner (CMP). CMP integrated navigation into the pipeline and also changed the global map representation to an egocentric latent map representation.~\cite{henriques2018mapnet} focused on extending the mapping aspect of~\cite{gupta2017cognitive} through the introduction of MapNet, which learns a ground-projection allocentric latent map of the explored environment. MapNet performs a brute-force localisation process at every time step; by doing so, temporal information is lost and irrelevant areas in the map are considered as viable localisation options. In contrast, our work uses this temporal information as a prior for localisation and updating the internal map.

\section{Embedded Memory Points Network}
\label{sec:empnet}



In Fig.~\ref{fig:system_overview}, a illustrative overview of our system is shown and a brief descriptive summary of our method is provided in the next subsection. Following this, we describe in more detail each core step of our framework. For the remainder of the paper, we use non-bold subscripts to represent matrices or scalars (depending on the context, \textit{i.e.} $R$), bold subscripts to represent vectors (\textit{i.e.} $\bm{q}$) and indexing into both is done using brackets (\textit{i.e.} $A[i,j]$ or $\bm{q}[i]$). Additionally, we refer to the central memory unit of our system, the Short-term Spatial Memory Module (SSMM), as two components, denoted as $\mathcal{M}_f$ and $\mathcal{M}_c$ which indicates the respective stored embeddings and their corresponding 3D points in the SSMM.

\subsection{System Overview}

An incoming RGB-D observation at time $t$, $x_t \in \mathbb{R}^{h\times w\times 4}$ of height $h$ and width $w$, is processed by a CNN (Section~\ref{subsec:point_embeds}) to produce the embeddings $h_{t,f} \in \mathbb{R}^{N_r \times n}$. Each embedding's corresponding locations in egocentric camera coordinates $h_{t,c} \in \mathbb{R}^{N_r \times 3}$ is obtained through projecting the depth information using the camera intrinsic matrix $K$. $h_{t,f}$ and $h_{t,c}$ represent the collectively generated point-embeddings. $N_r$ is the number point-embeddings generated and $n$ indicates the number of embedding channels. 


Computing pairwise distances between embeddings $h_{t,f}$ and $\mathcal{M}_{t-1,f} \in \mathbb{R}^{N_r b \times n}$, produces the distance map $\mathcal{D}_{t,f} \in \mathbb{R}^{N_r b \times N_r}$ (Section \ref{subsec:ssmm}); with $b$ denoting the buffer size of $\mathcal{M}$. The distance map $\mathcal{D}_{t,f}$ is converted into a Confidence Map $\mathcal{L}_{t,f} \in \mathbb{R}^{N_r b \times N_r}$ by applying a column-wise softmax operation and obtains the weight vector $\bm{\omega}_t \in \mathbb{R}^{N_r}$. This allows the system to optimise for the relative pose $T_{t} \in \mathbb{SE}(3)$, between the downsampled point cloud $h_{t,c}$ and their corresponding matches in $\mathcal{M}_{t,c} \in \mathbb{R}^{N_r b \times 3}$ in a weighted least squares formulation (Section \ref{subsec:wls}). 


Finally, an update step is performed by populating $\mathcal{M}_{t-1,f}$ with $h_{t,f}$ and transforming the downsampled point cloud $h_{t,c}$ in egocentric coordinate frame to $\mathcal{M}_{t-1,c}$'s coordinate frame by applying the estimated pose $T_{t}$ on $h_{t,c}$ and populating $\mathcal{M}_{t-1,c} \in \mathbb{R}^{N_r b \times 3}$, resulting in an updated $\mathcal{M}_{t,f}$ and $\mathcal{M}_{t,c}$. For the rest of the paper, for reducing clutter the time subscript $t$ will be omitted unless specified otherwise.

\subsection{Extracting Point Embeddings}
\label{subsec:point_embeds}

To extract point-embeddings from observations, we use a CNN architecture which receives an RGB-D input $x \in \mathbb{R}^{h \times w \times 4}$ and produces a tensor $x' \in \mathbb{R}^{h' \times w' \times n}$ where $h' < h, w' < w$, and $n$ is the channel length of each embedding. At this stage, we need to associate an embedding in $\mathbf{x'}[i,j,.]$ with the 3D point it represents. This is accomplished in the following manner: first, the given depth map $D \in \mathbb{R}^{h \times w}$ is resized to $D' \in \mathbb{R}^{h' \times w'}$ such that it matches the spatial dimensions of $x'$. In this case, a traditional bilinear downsampling approach was found to be sufficient. Next, we compute the 3D location of each entry in $D'[i,j]$ in egocentric camera coordinates using the known camera intrinsic matrix $K$ and the downsampled depth map $D'$ as shown below:
\begin{equation}
P_{c}[i,j,k] = D'[i,j]K^{-1}\begin{bmatrix} i, j, 1 \end{bmatrix}^{\top}
\label{eq:3d_proection}
\end{equation}
where $P_{c} \in \mathbb{R}^{h' \times w' \times 3}$. An entry in $P_{c}[i,j,k]$ is a 3D point in an egocentric camera coordinate frame corresponding to the depth map entry $D'[i,j]$. Rearranging $x' \in \mathbb{R}^{h' \times w' \times n}$ to $h_f \in \mathbb{R}^{N_r \times n}$ with $N_r=h'w'$ and similarly $P_c$ to $h_c \in \mathbb{R}^{N_r \times 3}$ yields two initial inputs to the next component, which localises both $h_f$, $h_c$ against $\mathcal{M}_{f}, \mathcal{M}_{c}$ that contains previously stored embeddings and 3D points in $\mathcal{M}_c$'s coordinate frame.

\subsection{Short-term Spatial Memory Localisation}
\label{subsec:ssmm}

We require the output of the CNN, $h_f$, to endow the framework with embeddings that can coherently match another set of point embeddings contained in $\mathcal{M}_f$. For this, we develop a loss function stemming from the ICP algorithm~\cite{arun1987least}. Finding the relative pose between two sets of point clouds requires finding the matching correspondences between them. Typically, two key difficulties emerge from this task: points not having any correspondences (due to partial overlaps), and points having correspondences which have more certainty than others. Here, we formulate the optimisation problem to address both issues using a unified weighting approach.

\begin{figure}[t]
\centering
\includegraphics[width=0.45\textwidth]{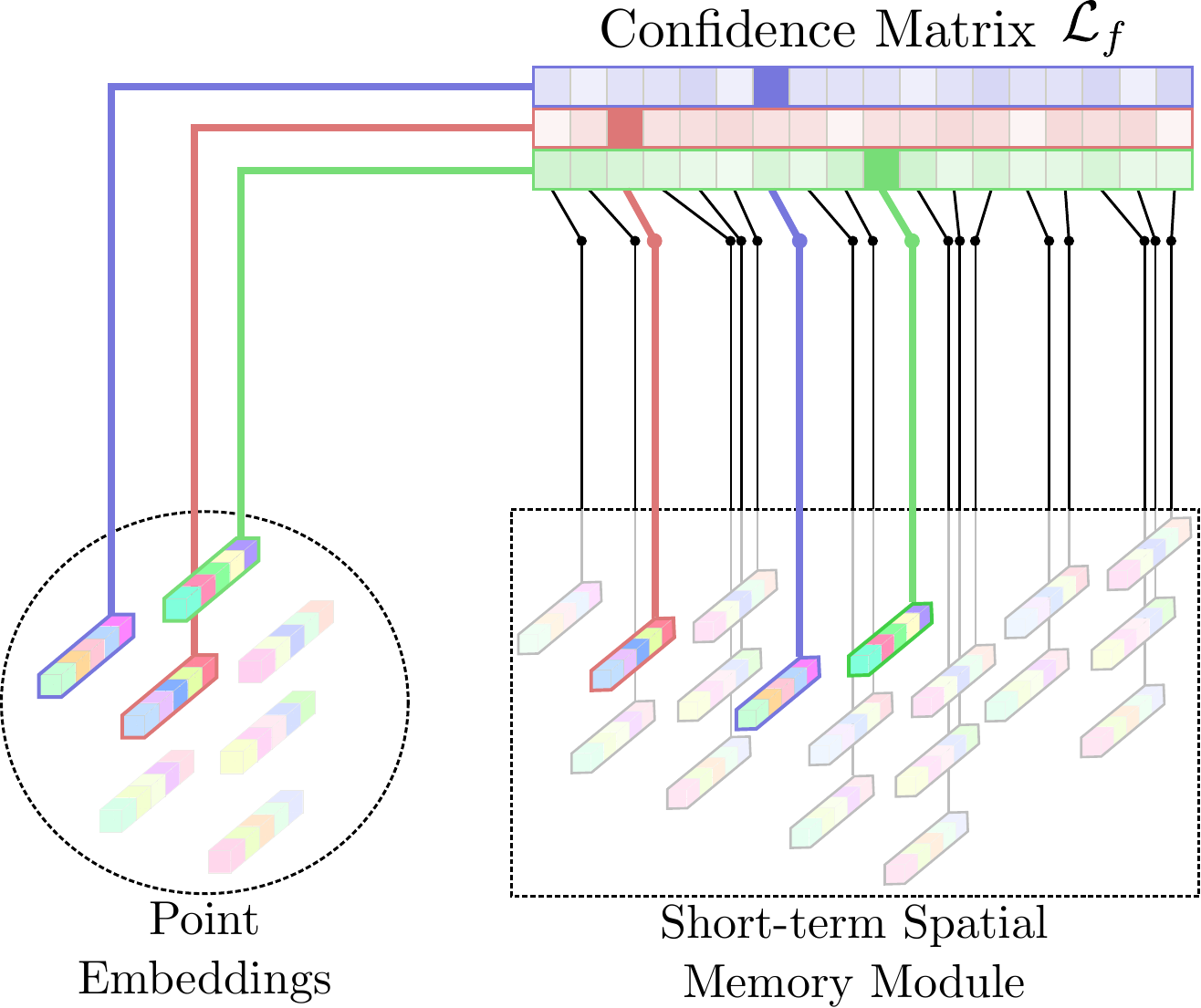}
\caption{The confidence matrix is constructed by densely computing a distance between embedded points of a current observation and the stored embedded memory points in the Short-term Spatial Memory Module. A darker colour indicates a higher confidence value in the confidence matrix.}
\label{fig:confidence_matrix}
\end{figure}

For an incoming observation $x$, we extract point-embeddings $h_f$ in the manner described in Section \ref{subsec:point_embeds}. We define the following operation as taking pairwise distances between the embeddings $\mathcal{M}_f$ and $h_f$:
\begin{equation}
\mathcal{D}_{f}[i,j] = d_{\phi}(\bm{\mathcal{M}}_{f}[i,.], \bm{h}_f[j,.])
\label{eq:embed_distances}
\end{equation}
Where $\bm{\mathcal{M}}_f[i,.], \bm{h}_f[j,.] \in \mathbb{R}^n$ are embedding row vectors, $\mathcal{D}_f \in \mathbb{R}^{N_r b \times N_r}$ is the pairwise distances matrix for the embeddings and $d_{\phi}$ is a distance metric on the embedding space. 
Reformulating $\mathcal{D}_{f}$ by applying the softmax operation yields:
\begin{equation}
\mathcal{L}_{f}[i,j] = \frac{e^{-\mathcal{D}_{f}[i,j]}}{\sum_{i'=1}^{N_r b}e^{-\mathcal{D}_{f}[i',j]}}
\label{eq:embed_softmax}
\end{equation}
Where $\mathcal{L}_{f} \in \mathbb{R}^{N_r b \times N_r}$ is the confidence matrix between embeddings in $\mathcal{M}_{f}$ and $h_f$. Note that a single column vector $\mathbf{\mathcal{L}}_{f}[.,j]$ is a confidence vector between the point $\bm{h}_c[j,.]$ and the entire set of points in $\mathcal{M}_{c}$ (Fig.~\ref{fig:confidence_matrix} illustrates this operation). 

For optimising the confidence matrix $\mathcal{L}_f$ towards a ground-truth confidence matrix $\mathcal{L}_{gt}$, we define our loss function as the cross-entropy loss:
\begin{equation}
loss_{c} = -\frac{1}{N_r}\sum_{j=1}^{N_r}\sum_{i=1}^{N_r b}\mathcal{L}_{gt}[i,j]\log\mathcal{L}_f[i,j]
\label{eq:loss}
\end{equation}
The ground-truth confidence matrix $\mathcal{L}_{gt} \in \mathbb{R}^{N_r b \times N_r}$ is computed using a procedure similar to the one outlined above; more explicitly, we define $\mathcal{M}_{gt} \in \mathbb{R}^{N_r b \times 3}$ to be a sequence of point clouds, aligned using the ground-truth poses to a shared coordinate frame.
For an incoming ground-truth aligned point cloud $h_{gt}=T_{gt}h_c$ at time $t$ which follows point sequences stored in $\mathcal{M}_{gt}$, the ground-truth pairwise distances are computed as follows:
\begin{equation}
\mathcal{D}_{gt}[i,j] = \norm{\bm{\mathcal{M}}_{gt}[i,.] - \bm{h}_{gt}[j,.]}_{2}
\label{eq:gt_distances}
\end{equation}
Where $\bm{\mathcal{M}}_{gt}[i,.], \bm{h}_{gt}[j,.] \in \mathbb{R}^{3}$ are row vectors. 
A property of $\mathcal{D}_{gt}$ is that points which are close enough will have a small distance value whilst points which do not have a match (\textit{i.e.} are in a non-overlapping region) will have a large distance. 
This can be exploited in a way which will amplify both matching and non-matching cases, where a probability `$1$' is assigned to matches and a `$0$' to non-matches. Similar to Eq.~\ref{eq:embed_softmax}, we reformulate matrix $\mathcal{D}_{gt}$:
\begin{equation}
\mathcal{L}_{gt}[i,j] = \frac{e^{-\tau \mathcal{D}_{gt}[i,j]}}{\sum_{i'=1}^{N_r b}e^{-\tau \mathcal{D}_{gt}[i',j]}}
\label{eq:gt_softmax}
\end{equation}
The temperature coefficient $\tau$ controls the amplification of distance correspondences and is a hyper-parameter. Finally, we note the operations discussed in this section are naturally parallelised and can be computed efficiently using modern GPU architectures. 

\subsection{Best-fitting of Weighted Correspondences}
\label{subsec:wls}
In the previous sections, we formulated a loss which optimises the embeddings $h_f$ to follow the 3D closest point criteria. This allows for the recovery of a matrix $\mathcal{L}_{f}$, which contains confidence values between a point in $h_c$ and $\mathcal{M}_c$. These confidence values represent weights that can be used for applying a weighted best-fit algorithm. The weights are obtained as follows:
\begin{equation}
\omega[j] = \max_{i'}\mathcal{L}_{f}[i',j]
\label{eq:max_weights}
\end{equation}
and respectively, the point index in $\mathcal{M}_c$ corresponding to point $h_c[j,.]$:
\begin{equation}
c[j] = \argmax_{i'}\mathcal{L}_{f}[i',j]
\label{eq:argmax_weights}
\end{equation}
Where $\bm{c} \in \mathbb{R}^{N_r}$ is the indexing vector for aligning the correspondences in $\mathcal{M}_c$ to those in $h_c$. The weights in vector $\bm{\omega} \in \mathbb{R}^{N_r}$ are the respective confidences for those matches. Computing the relative-pose between the point cloud $h_c$ in egocentric coordinates and the points in $\mathcal{M}_{c}$, with its respective coordinate frame, can be estimated using a weighted best-fit approach.

\paragraph{Weighted Best-fitting}
Given two sets of point clouds $p \in \mathbb{R}^{M \times 3}$, their correspondences $q \in \mathbb{R}^{M \times 3}$ and weight vector $\bm{\omega} \in \mathbb{R}^{M}$. The rigid-transform can be computed in a closed form and is optimal in a weighted least-squares sense (proof in~\cite{hartley2003multiple}). Formally, we solve:
\begin{equation}
R,\mathbf{t} = \argmin_{R \in SO(3), \mathbf{t} \in \mathbb{R}^{3}} \sum_{\ell=1}^{M}\omega[\ell]\norm{\mathbf{q}[\ell,.] - (R\mathbf{p}[\ell,.] - \mathbf{t})}_{2}
\label{eq:WLS}
\end{equation}
For obtaining $R, \mathbf{t}$ we initially compute:
\begin{equation}
\Bar{\mathbf{p}} = \frac{\sum_{\ell=1}^{M}\omega[\ell]\mathbf{p}[\ell,.]}{\sum_{\ell=1}^{M}\omega[\ell]}, \quad \Bar{\mathbf{q}} = \frac{\sum_{\ell=1}^{M}\omega[\ell]\mathbf{q}[\ell,.]}{\sum_{\ell=1}^{M}\omega[\ell]}
\label{eq:weighted_means}
\end{equation}
With $\Bar{\mathbf{p}}, \Bar{\mathbf{q}} \in \mathbb{R}^3$ being the weighted average centroids of $p, q$ respectively. By subtracting each weighted centroid from its respective point cloud we get:
\begin{equation}
\mathbf{\hat{p}}[\ell,.] = \mathbf{p}[\ell,.] - \Bar{\mathbf{p}}, \quad \mathbf{\hat{q}}[\ell,.] = \mathbf{q}[\ell,.] - \Bar{\mathbf{q}}
\label{eq:centered_pcs}
\end{equation}
Finally, by defining $\Omega = diag(\mathbf{\omega})$ and applying SVD decomposition such that: $U\Sigma V^{\top} = \hat{p}^{\top}\Omega\hat{q}$
, the rotation matrix $R$ is computed as:
\begin{equation}
R = V diag(1, 1, \det(VU^{\top}))U^{\top}
\label{eq:rotation_mat}
\end{equation}
and the translation vector $\mathbf{t}$:
\begin{equation}
\mathbf{t} = \mathbf{\Bar{q}} - R\mathbf{\Bar{p}}
\label{eq:translation_vec}
\end{equation}
\\
By performing the weighted best-fit procedure outlined above, we obtain the relative pose $T$ between $h_c$ in egocentric coordinate frame and $\mathcal{M}_c$ in its respective coordinate frame. In the last step of our framework, $h_c$ is transformed using the estimated pose yielding $h_{c}^{'} = Th_{c}^{\top}$, where populating $\mathcal{M}_c$ with $h_{c}^{'}$ and $\mathcal{M}_f$ with $h_f$ need not be in a particular order, as both are unstructured. \\

A straight-forward extension to this approach is imposing pose regularisation on the loss developed in Section~\ref{subsec:ssmm}. This is achieved through projecting $\mathcal{M}_{c}$ onto the confidence matrix $\mathcal{L}_{f}$:
\begin{equation}
\bar{\mathcal{M}}_{c} = \mathcal{L}_{f}^{\top}\mathcal{M}_{c}
\label{eq:weighted_matches}
\end{equation}
with $\bar{\mathcal{M}}_{c} \in \mathbb{R}^{N_r \times 3}$ being the correspondences of $h_c$. The best-fit approach (without weights $\bm{\omega}$) from Eq.~\ref{eq:WLS} is applied to obtain the rotation matrix $R$ and translation vector $\mathbf{t}$. This alternative formulation also makes our method fully differentiable. Finally, we modify the loss given in Eq.~\ref{eq:loss} by adding two regularisation terms:
\begin{equation}
Loss = loss_{c} + \lambda_{R}loss_{R} + \lambda_{t}loss_{t}
\label{eq:loss_reg}
\end{equation}
both $loss_{R}$ and $loss_{t}$ are formulated as in~\cite{kendall2015posenet}. In summary, both approaches formulated differ by how the confidence matrix $\mathcal{L}_{f}$ is used to compute the pose (Eq.~\ref{eq:loss} and Eq.~\ref{eq:loss_reg}).


\section{Experiments}
\label{sec:experiments}
To compare our models qualitatively and quantitatively, we perform experiments on two challenging benchmarks: a synthetic environment (VIZDoom~\cite{kempka2016vizdoom}) and a real world indoor environment (Active Vision Dataset~\cite{ammirato2017dataset}). We evaluate two variants of our model: EMP-Net which optimises Eq.~\ref{eq:loss} and EMP-Net-Pose which optimises Eq.~\ref{eq:loss_reg}. For the Doom dataset, we compared our framework against DeepVO~\cite{wang2017deepvo} and MapNet~\cite{henriques2018mapnet}, both which maintain an internal representation of previously seen observations. Additionally, we compared our models against a recent state-of-the-art frame-to-frame approach ENG~\cite{dharmasiri2018eng}. For the AVD dataset, we also compare with a mature classic SLAM baseline, an RGB-D implementation of ORB-SLAM2~\cite{mur2015orb}.

\paragraph{Network Architecture}
For extracting feature embeddings, we employ a U-Net architecture~\cite{ronneberger2015u}. We initialised our network weights using the initialisation scheme detailed in~\cite{he2015delving}. The U-Net uses an encoder-decoder setup where the encoder consists of three encoder blocks separated by max pooling layers and the decoder consists of two decoder blocks. Each block in the encoder consists of two sequences of convolution layers comprising of $3\times3$ filters, followed by Batch Normalisation~\cite{ioffe2015batch} and ReLU Activation~\cite{nair2010rectified}. Each block in the decoder consists of a transposed convolution layer followed by Batch Normalisation~\cite{ioffe2015batch}, ReLU Activation~\cite{nair2010rectified} followed by a convolution layer. The transposed convolution layer upsamples the input using a stride 2 deconvolution, the output of which is concatenated to its matching output from the encoder block.

\paragraph{Training Settings}
For both synthetic and real experiments, our EMP-Net model is trained with a batch size of 16, where every instance within a batch is a sequence of 5 consecutive frames resized to $120 \times 160$. The RGB and depth data were scaled to between $[0,1]$. The buffer size of the SSMM is $b=4$ and the number of extracted point-embeddings is $N_r=4800$. The temperature parameter was tested with values between $\tau=[10^{3},10^{6}]$, where we found the model to be fairly invariant to this value. In all of the experiments shown $\tau=10^{5}$. The embedding distance function is defined as the $L2$ distance, $d_{\phi}=\norm{a-b}_2$. $\lambda_t=0.02$ and $\lambda_R=5$ are chosen to maintain the same ratio as in~\cite{kendall2015posenet}. We use the ADAM optimiser~\cite{kingma2014adam}, using the default first and second moment terms of 0.9 and 0.999 values respectively. We use a learning rate of $10^{-3}$ and train for 10 epochs.

\paragraph{Error Metrics}
Across both datasets, we quantitatively benchmark against baselines using two error metrics. We measure the Average Position Error (APE), which denotes the average Euclidean distance between predicted position of the agent and a corresponding ground-truth position. Additionally, we inspect the Average Trajectory Error (ATE) which describes the minimum RMS error in position between a translated and rotated predicted trajectory \textit{w.r.t} a ground-truth trajectory. Thus, for longer sequences the APE will naturally be worse as it does not correct for drifts occurring over time. Similar to~\cite{henriques2018mapnet}, we measure both short-term APE over 5 observation frames (APE-5) as well as long term APE (APE-50) and ATE (ATE-50) over 50 observation frames.

\subsection{Synthetic 3D Data}
\label{ssec:exp_doom_ape}
We used VIZDoom to record human players performing 4 speed-runs of the game with in-game sprites and enemies turned off. Despite VIZDoom being a synthetic environment it provides rich and complex visual scenarios that emulate the difficulties encountered in real world settings. The captured recordings include RGB-D and camera pose data, which correspond to 120k sequences. Training sequences are composed of 5 frames, sampled every second frame of recorded video. For testing, we randomly select sequences of 50 consecutive frames and remove those from the training set construction.


\begin{figure}[t]
\centering
\includegraphics[width=0.4\textwidth]{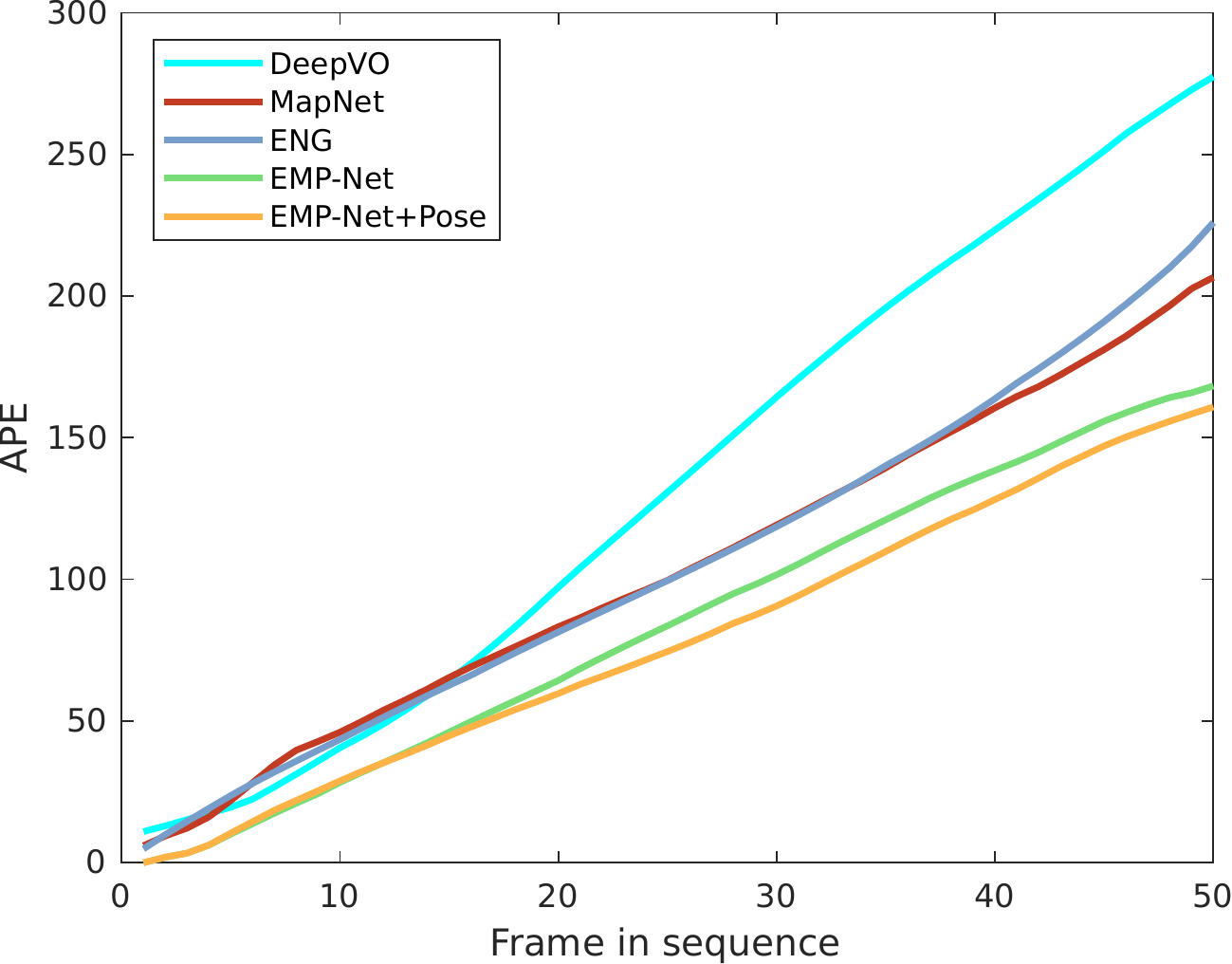}
\caption{Average Positional Error (APE) over different sequence lengths (5-50 frames) on the Doom dataset}
\label{fig:doom_ape_plots}
\vspace{-1.0em}
\end{figure}
\begin{table}[ht]
\small
\begin{center}
\setlength\tabcolsep{0.40cm}
\begin{tabular}{c c c c c}
\specialrule{.2em}{.1em}{.1em}
\thead{Doom data~\cite{kempka2016vizdoom}} &\thead{APE-5} &\thead{APE-50} &\thead{ATE-50} \\
\hline
DeepVO~\cite{wang2017deepvo} & 19.56 & 277.4 & 111 \\
MapNet~\cite{henriques2018mapnet} & 21.98 & 206.6 & 76 \\
ENG~\cite{dharmasiri2018eng} & 23.71 & 225.9 & 105 \\
EMP-Net (Ours) & \textbf{10.10} & 168.3 & 68 \\
EMP-Net-Pose (Ours) & 10.45 & \textbf{160.9} & \textbf{59} \\
\hline
\end{tabular}
\caption{Average Position Error (APE) and Absolute Trajectory Error (ATE) on VIZDoom dataset.}
\label{tab:doom_ape_ate}
\vspace{1.0em}
\end{center}
\end{table}

\paragraph{Quantitative Results}
We measure APE across the test sequences on all the above mentioned models (Fig.~\ref{fig:doom_ape_plots}). Increasing the sequence length beyond a sequence length of 5 examines its ability to generalise beyond the length of the training sequences. Both DeepVO\cite{wang2017deepvo} and ENG\cite{dharmasiri2018eng} lack an internal map representation to localise against and similarly both methods suffer from a larger accumulated drift towards the end of the sequence. MapNet\cite{henriques2018mapnet} fairs better, although suffers from inaccuracies due to cell quantisation and false pose modalities which appear over longer periods of time. We note that both our non-regularised and regularised methods (EMP-Net and EMP-Net-Pose) significantly outperform the compared baselines across all sequence lengths. The two variants are similar in their performance, with a marginal improvement that is gained by the additional pose regularisation. This minor improvement is explained from the nature of the data. VIZDoom provides noiseless depth and pose information which correspond perfectly to each other, as both are obtained directly from the game engine. In other words, the provided ground-truth labels for regressing the confidence matrix contain all the necessary information regarding the pose.
In Table~\ref{tab:doom_ape_ate}, we show the APE across observation sequences of 5 frames and 50 frames, as well as the ATE across a 50 observation sequence.

\begin{figure}
\centering
\subfloat[]{%
\includegraphics[width=0.44\textwidth]{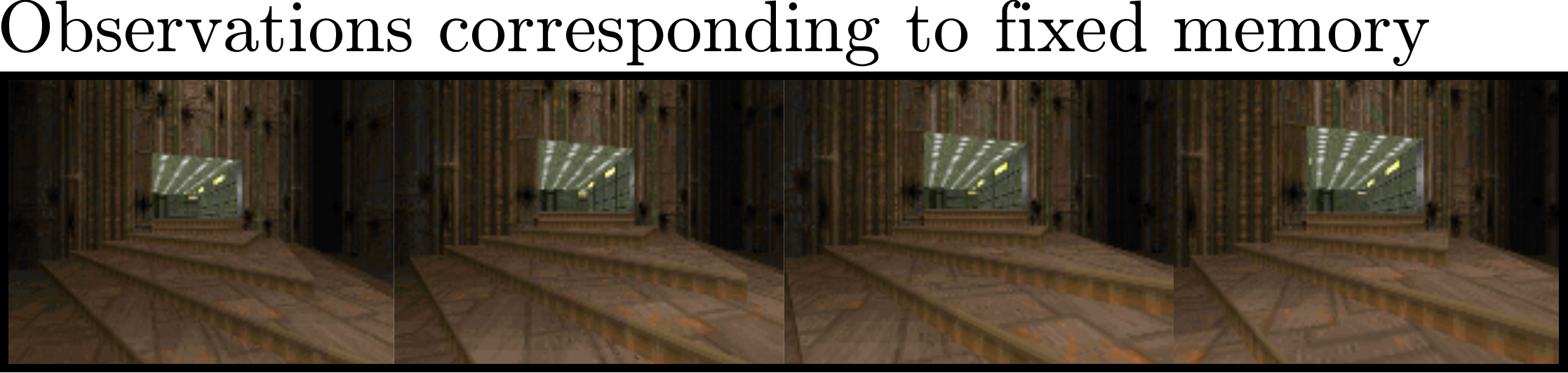}
\label{fig:doom_exp2_figures_fixed_memory}
} \\
\vspace{1.0em}
\rulesep
\subfloat[]{%
\includegraphics[width=0.44\textwidth]{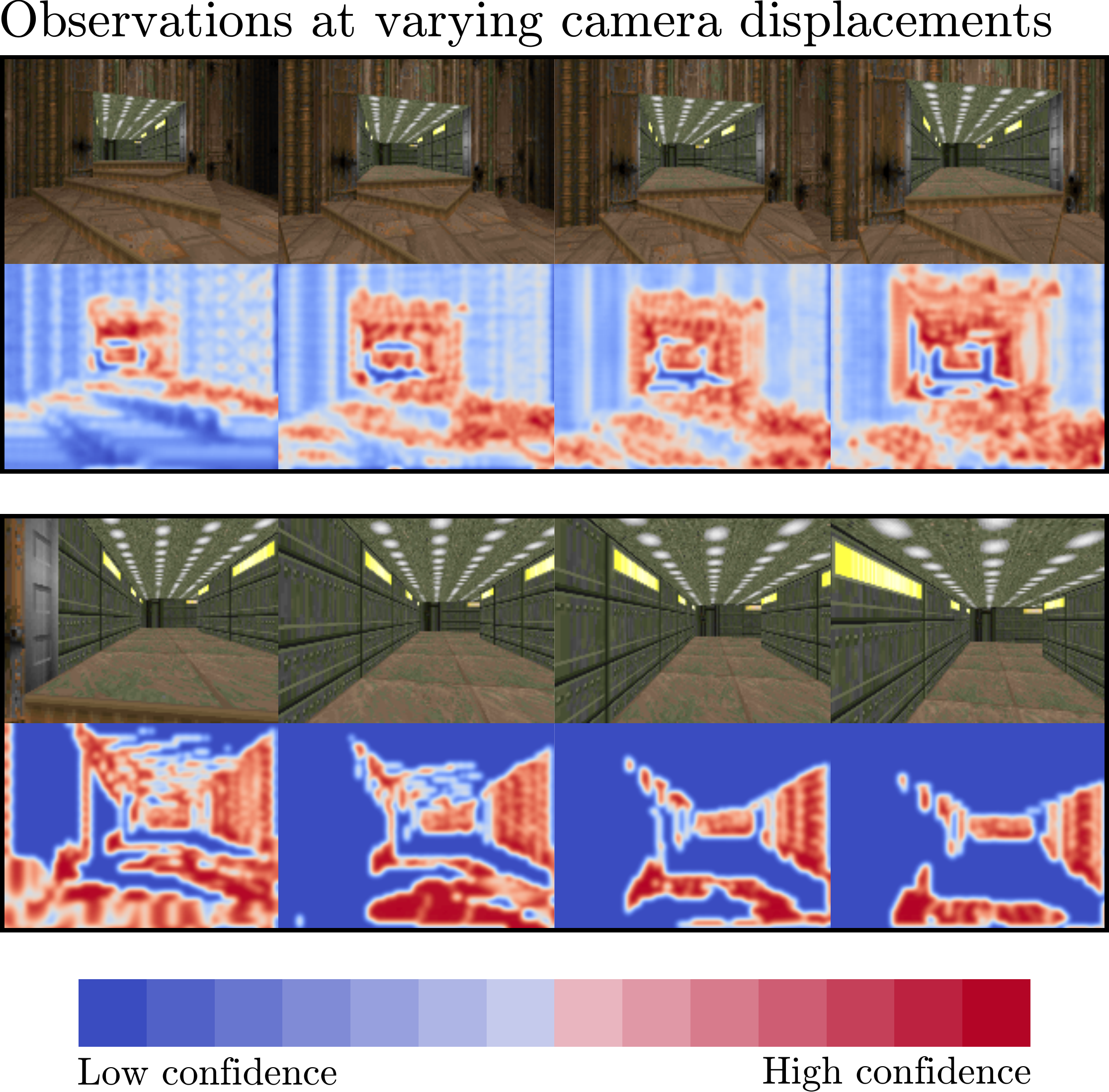}
\label{fig:doom_exp2_figures_new_sequences}
}
\caption{Qualitative results of Section~\ref{ssec:exp_analysis}. (a): Observations corresponding to the embedded memory points stored in EMP-Net's memory module. (b): Incoming observations which are processed and localised against the stored embedded memory. For distinguishable landmarks (\textit{i.e.} stairs and corridor entrance) confidence values are higher. In later frames, areas that are not visible in the past observations are correctly receiving near zero confidence.}
\label{fig:doom_exp2_fig}
\end{figure}

\subsection{Confidence Matrix Interpretation}
\label{ssec:exp_analysis}
\begin{figure}[t!]
\centering
\includegraphics[width=0.4\textwidth]{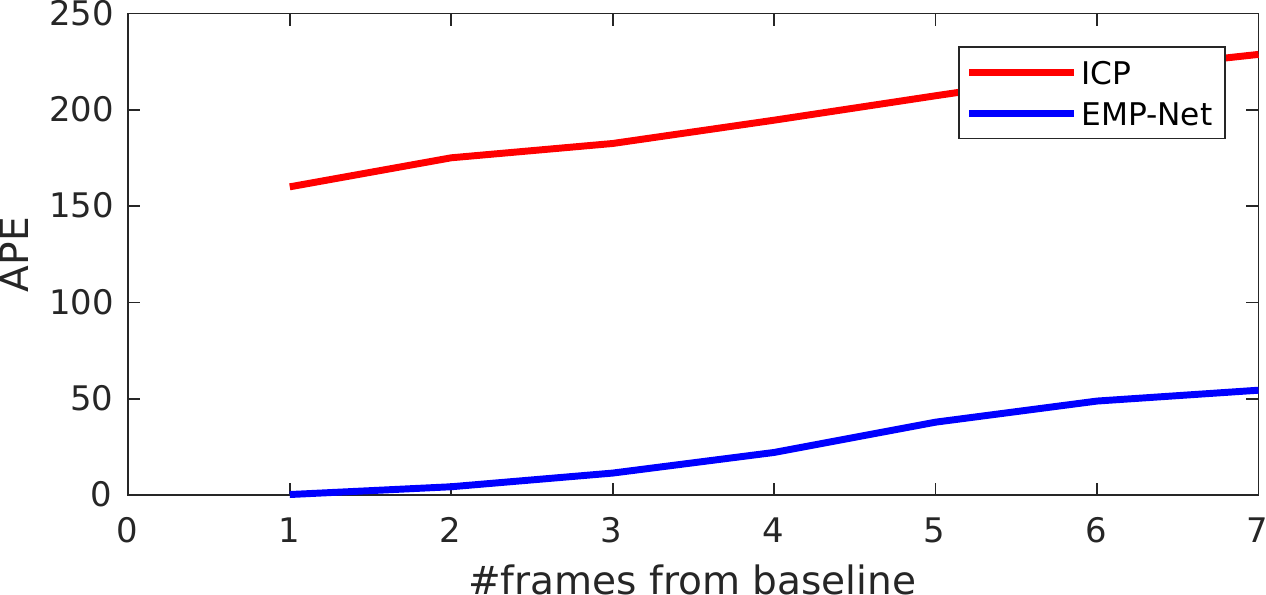} \\
\includegraphics[width=0.4\textwidth]{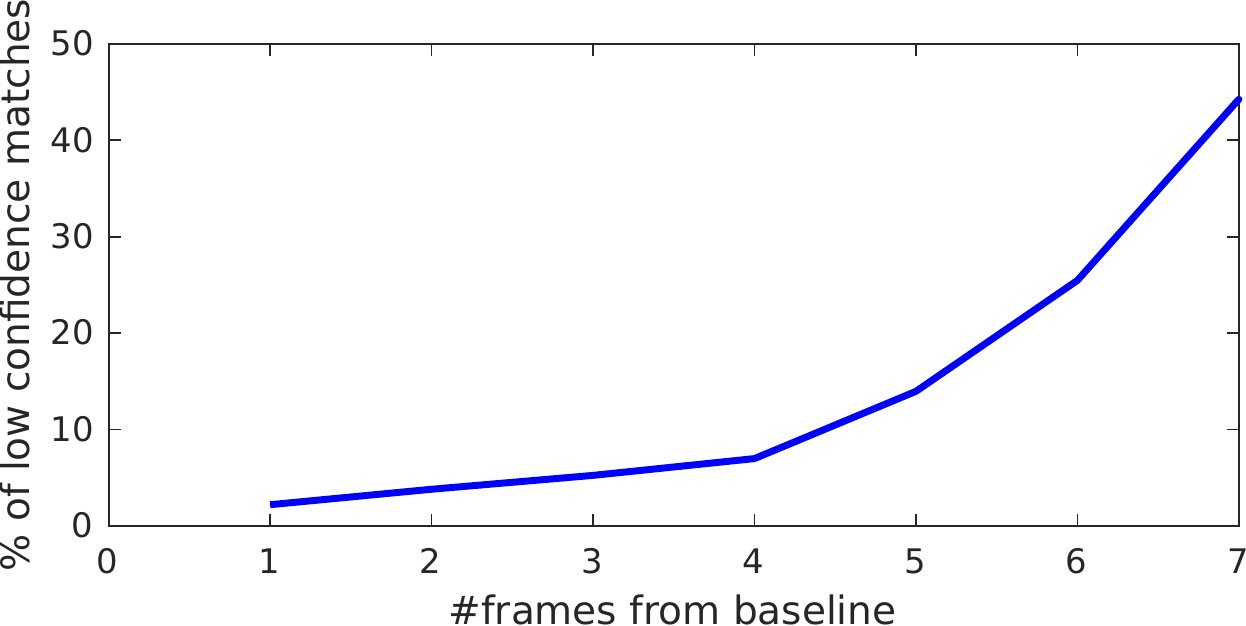}
\caption{\textbf{Top}: APE measured over increasing camera baselines on the Doom dataset. For reference we compare to traditional ICP algorithm\cite{hartley2003multiple}. \textbf{Bottom}: Percentage of low confidence matches over increasing baselines. In this case, low confidence indicates a confidence of less than 0.05. As the baseline increases, the agent becomes increasingly less confident.}
\label{fig:doom_exp2_plots}
\end{figure}

In this section, we provide additional insight about the inferred confidence matrix in EMP-Net. We run an experiment which allows the system to process 
test sequences and store point-embeddings up to the size of the buffer (\textit{i.e} four observation frames). Beyond this point, we discontinue storing point embeddings but continue to process the sequence by localising against the existing embeddings within the SSMM. This allows for simulation of large camera motions and assesses the robustness of the estimated confidence matrix across increasingly larger camera motions. The top figure in Fig.~\ref{fig:doom_exp2_plots} shows the APE of EMP-Net as the camera baseline grows. For reference, we plot the APE for a standard ICP~\cite{hartley2003multiple} point-to-point implementation. Note that while the APE value increases as we shift further away from the baseline, EMP-Net is demonstrably more robust to larger shifts in the camera baseline. The reduced performance on APE correlates with lower confidence of the system as evidenced by the bottom figure in Fig.~\ref{fig:doom_exp2_plots}.

In Fig.~\ref{fig:doom_exp2_figures_new_sequences}, we show confidence heat maps along with their corresponding observation frames computed against a fixed memory from observations in Fig.~\ref{fig:doom_exp2_figures_fixed_memory}. These confidence heat maps are obtained by reshaping the confidence weight vector $\bm{\omega}$ to the size of the downsampled observation frame ($60\times80$). Note that higher confidence is assigned towards landmarks with distinguishable features (\textit{e.g.} stairs, corridor entrance, \textit{etc.}), whilst lower confidence is assigned to low texture landmarks (\textit{e.g.} walls and floor). For frames with little overlap, the system assigns high confidence to landmarks that it is able to locate in its memory buffer (\textit{i.e.} visible in ``fixed memory" frames). At times, these confidences may be overestimated due to a lack of better correspondences from its memory.


\begin{figure*}[t]
\centering
\hspace{2.0em}
\includegraphics[width=0.45\textwidth]{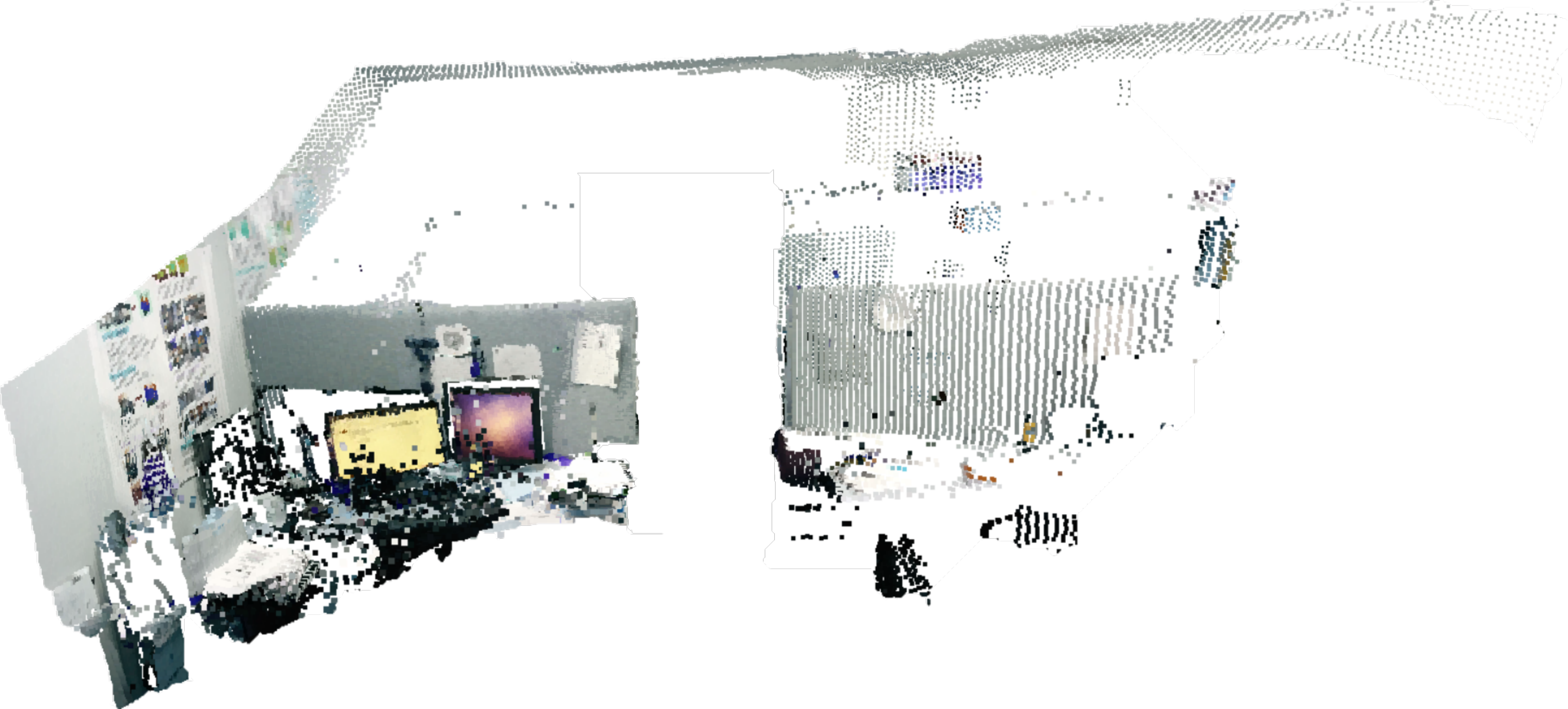}
\hspace{2.0em}
\includegraphics[width=0.45\textwidth]{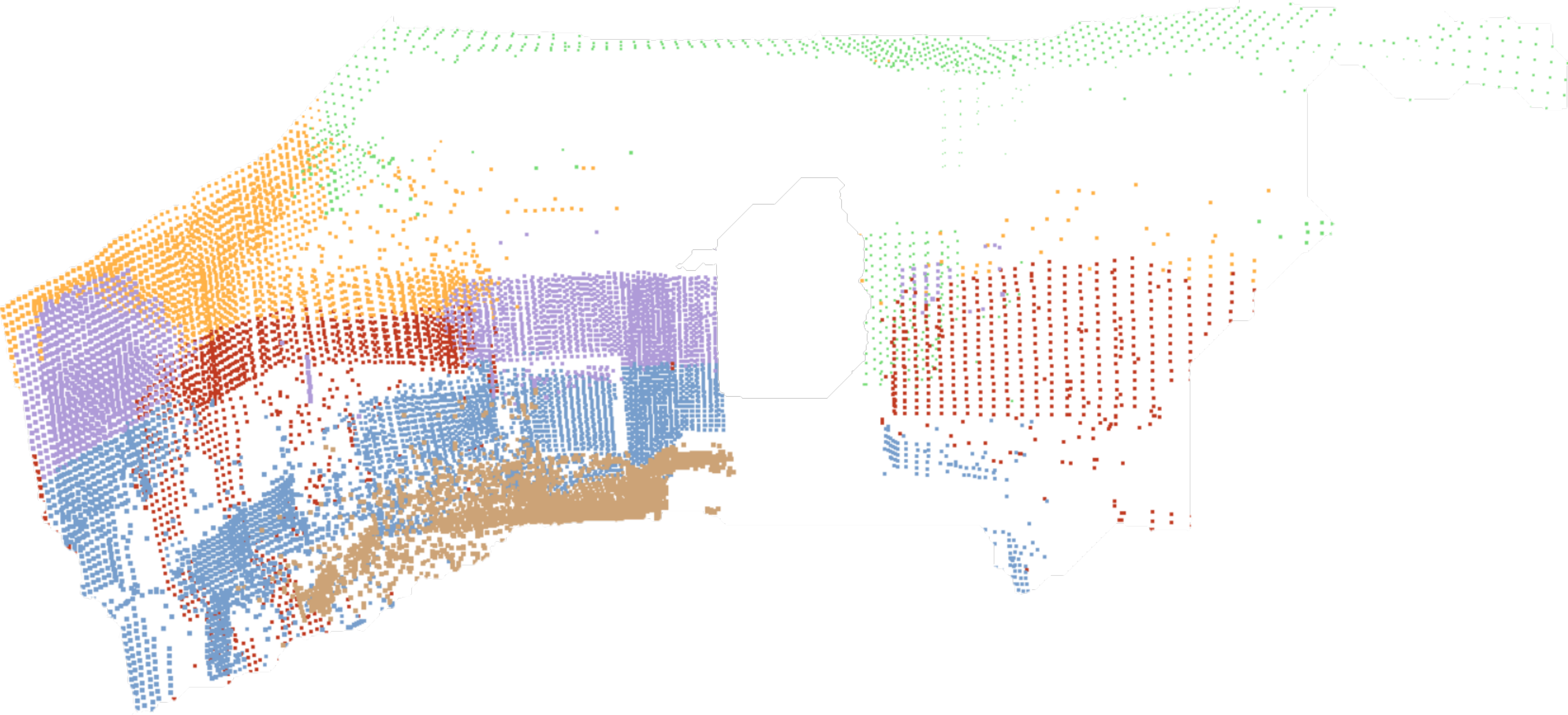}
\caption{Inspecting the point embedding values provides insight of the underlying operation of EMP-Net. \textbf{Left}: The ground-truth alignment of a sequence of 5 frames. \textbf{Right}: The corresponding downsampled point cloud in the \textit{SSMM}. The colors define cluster centers of the embeddings. We observe a mixture of spatial and semantic segmentation that is learned without explicit supervision.}
\label{fig:avd_segment}
\end{figure*}

\subsection{Real World Data}
\begin{figure}
\centering
\includegraphics[width=0.4\textwidth]{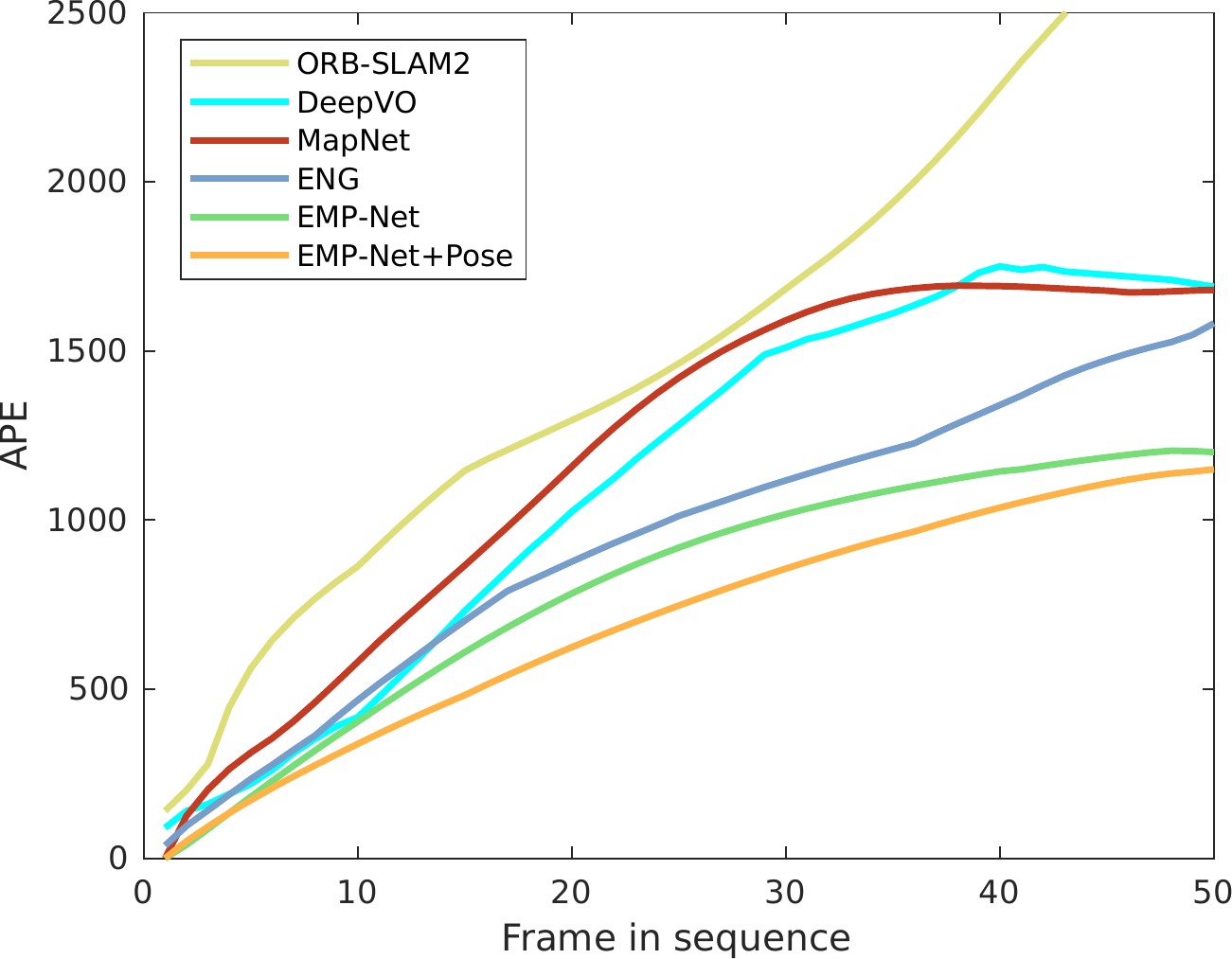}
\caption{Average Positonal Error (APE) over different sequence lengths (5-50 frames) on the Active Vision Dataset}
\label{fig:avd_ape_plots}
\vspace{-1.0em}
\end{figure}

For our real world data experiments, we use the Active Vision Dataset (AVD)~\cite{ammirato2017dataset}. This dataset consists of RGB-D images across 19 indoor scenes. Images are captured by a robotic platform which traverses a 2D grid with translation steps size of 30cm and 30$\degree$ in rotation. For generating robot navigation trajectories, the captured images can be arbitrarily combined. Similar to \cite{henriques2018mapnet}, for training, we sampled 200,000 random trajectories, each consisting of 5 frames, where the trajectory was chosen using the shortest path between 2 randomly selected locations from 18 out of the 19 provided scenes. For testing, we sampled 50 random trajectories, each consisting of 50 frames, where the trajectory was chosen from the unseen test scene.
\paragraph{Quantitative and Qualitative Results}
We measure APE across the test sequences and show results in Fig.~\ref{fig:avd_ape_plots}. Once again, we increase the sequence length for testing to sequences beyond 5 observation frames to evaluate the ability to generalise beyond training sequence length. Both the non-regularised and regularised methods of EMP-Net significantly outperform the compared baselines across all sequence lengths. In this case, unlike the VIZDoom environment, the use of real world data is accompanied with noisy sensory measurements. Consequently, EMP-Net-Pose is observably more robust than its non-regularised version.
In Table~\ref{tab:avd_ape_ate}, we show the APE across observation sequences of 5 frames and 50 frames, as well as the ATE across a 50 observation sequence for the test AVD dataset. \\
In Fig. \ref{fig:avd_segment} we provide additional insight on interpreting the information contained in the learned embeddings of EMP-Net. A snapshot of the SSMM can be seen in Fig. \ref{fig:avd_segment} (Right), where we show a downsampled point cloud stored in the SSMM with the inferred aligning. Each point in the SSMM has a corresponding embedding vector. The colour assigned to each point is a cluster centroid colour code that was obtained by performing a \textit{k-means} clustering over the embeddings. A mixture of spatial and semantic segmentation can be observed. For reference, Fig.~\ref{fig:avd_segment} (Left) is the ground-truth alignment of the point clouds obtained at the original resolution with their corresponding RGB values. For additional qualitative results on the AVD dataset, please refer to our supplementary video material.

\label{ssec:real_data}
\begin{table}[ht]
\small
\begin{center}
\setlength\tabcolsep{0.20cm}
\begin{tabular}{c c c c c}
\specialrule{.2em}{.1em}{.1em}
\thead{AVD data~\cite{ammirato2017dataset}} &\thead{APE-5} &\thead{APE-50} &\thead{ATE-50} \\
\hline
ORB-SLAM2 (RGB-D)~\cite{mur2015orb} & 432  & 3090 & 794 \\
DeepVO~\cite{wang2017deepvo} & 220.0 & 1690 & 741 \\
MapNet~\cite{henriques2018mapnet} & 312.3 & 1680 & 601 \\
ENG~\cite{dharmasiri2018eng} & 234.3 & 1582 & 757 \\
EMP-Net (Ours) & 181.6 & 1201 & 381 \\
EMP-Net-Pose (Ours) & \textbf{171.8} & \textbf{1150} & \textbf{360} \\
\hline
\end{tabular}
\caption{Average Position Error (APE) and Absolute Trajectory Error (ATE) on the Active Vision Dataset.}
\label{tab:avd_ape_ate}
\vspace{1.0em}
\end{center}
\end{table}

\section{Future Work}
\label{sec:future_work}
In future work, we look towards extending EMP-Net to larger navigation problems by addressing the linear complexity growth of computing the correspondence matrix (\textit{i.e.} large buffer sizes). Extensions worth pursuing for reducing this complexity are non-dense methods for generating correspondences by using approximate nearest neighbour search like methods or formulating the vocabulary tree~\cite{nister2006scalable} so it can be integrated within modern deep learning frameworks.


{\small
\bibliographystyle{ieee}
\bibliography{paper_final.bib}
}

\end{document}